\documentclass[10pt,twocolumn,letterpaper]{article}

\usepackage{cvpr}              

\usepackage{graphicx}
\usepackage{amsmath}
\usepackage{amssymb}
\usepackage{booktabs}
\usepackage{verbatim}
\usepackage{siunitx}
\usepackage{array}

\usepackage[pagebackref,breaklinks,colorlinks]{hyperref}

\usepackage[capitalize]{cleveref}
\crefname{section}{Sec.}{Secs.}
\Crefname{section}{Section}{Sections}
\Crefname{table}{Table}{Tables}
\crefname{table}{Tab.}{Tabs.}

\def\cvprPaperID{*****} 
\def\confName{CVPR}
\def\confYear{2024}

\newcommand{\lcalib}{L^{\mathrm{calib}}}
\newcommand{\lsimple}{L^{\mathrm{simple}}}

\newcommand{\ocut}{\omega_{\mathrm{cut}}}

\newcommand{\Tcut}{T_{\mathrm{cut}}}
\newcommand{\noff}{N_{\mathrm{OFF}}}
\newcommand{\non}{N_{\mathrm{ON}}}

\newcommand{\nrange}{N_{\mathrm{range}}}
\newcommand{\coff}{C_{\mathrm{OFF}}}
\newcommand{\con}{C_{\mathrm{ON}}}

\newcommand{\Coffi}{\coff^{(i)}}
\newcommand{\Coni}{\con^{(i)}}
\newcommand{\Conip}{\con^{\prime(i)}}
\newcommand{\Coffip}{\coff^{\prime(i)}}
\newcommand{\Cpk}{C_{p_k}}
\newcommand{\Cpki}{C_{p_k}^{(i)}}
\newcommand{\Ci}{C^{(i)}}
\newcommand{\Cip}{C^{\prime(i)}}
\newcommand{\Non}{n_{\mathrm{ON}}}
\newcommand{\Noff}{n_{\mathrm{OFF}}}
\newcommand{\Ntot}{n_{\mathrm{tot}}}
\newcommand{\Npix}{n_{\mathrm{pix}}}
\newcommand{\Noffi}{\Noff^{(i)}}
\newcommand{\Noni}{\Non^{(i)}}
\newcommand{\Ntoti}{\Ntot^{(i)}}
\newcommand{\Ntotbar}{\bar{n}_\mathrm{tot}}
\newcommand{\Nap}{n_\mathrm{pix}^\mathrm{(act)}}
\newcommand{\Nat}{n_\mathrm{tiles}^\mathrm{(act)}}
\newcommand{\deltaL}{\Delta L}
\newcommand{\absdeltaL}{|\deltaL|}

\newcommand{\deltaLki}{\deltaL_k^{(i)}}
\newcommand{\pk}{p_k}
\newcommand{\pbark}{\bar{p}_k}
\newcommand{\Ldet}{L^\mathrm{\scalebox{0.75}{det}}}
\newcommand{\Lfibar}{L^\mathrm{\scalebox{0.5}{FIBAR}}}
\newcommand{\deltaLdet}{\Delta \Ldet}
\newcommand{\alphacut}{\alpha_\mathrm{cut}}
\newcommand{\betacut}{\beta_\mathrm{cut}}
\newcommand{\rfill}{r^\mathrm{(fill)}}
\newcommand{\qtarg}{q^\mathrm{(targ)}}

\newcolumntype{P}[1]{>{\centering\arraybackslash}m{#1}}

\begin{document}
\pdfsuppresswarningpagegroup=1
\title{Filter-Based Reconstruction of Images from Events}
\author{Bernd Pfrommer\\Event Vision Research LLC\\{\tt\small bernd.pfrommer@eventvisionresearch.com}}
\maketitle

\begin{abstract}
  Reconstructing an intensity image from the events of a moving event camera
  is a challenging task that is typically approached with neural networks deployed
  on graphics processing units. This paper presents a much simpler,
  {\bf FI}lter {\bf B}ased {\bf A}synchronous {\bf R}econstruction method ({\bf FIBAR}).
  First, intensity changes signaled by events are integrated with a temporal digital IIR filter.
  To reduce reconstruction noise, stale pixels are detected by a novel
  algorithm that regulates a window of recently updated pixels.
  Arguing that for a moving camera, the absence of events at a pixel location likely
  implies a low image gradient, stale pixels are then blurred with a Gaussian filter.
  In contrast to most existing methods, FIBAR is asynchronous and permits image read-out
  at an arbitrary time.
  It runs on a modern laptop CPU at about 42(140) million events/s with (without)
  spatial filtering enabled.
  A few simple qualitative experiments are presented that show the difference in image
  reconstruction between FIBAR and a neural network-based approach (FireNet).
  FIBAR's reconstruction is noisier than neural network-based methods and suffers from ghost images.
  However, it is sufficient for certain tasks such as the detection of fiducial markers.
  Code is available at
  \url{https://github.com/ros-event-camera/event_image_reconstruction_fibar}.
\end{abstract}

\section{Introduction}
\label{sec:introduction}
Event-based cameras~\cite{chakravarthi2024recenteventcamerainnovations}
imitate a biological retina by producing a signal whenever a pixel's
brightness\footnote{Following \cite{gallego_delbrueck_orchard},
brightness here refers to the logarithm of the photo current, which is closely related
to the logarithm of the illumination intensity.}
has increased (ON event) or decreased (OFF event) by more than a certain contrast threshold.
While event cameras offer low latency and high dynamic range, their adoption has been
tepid, because unlike traditional cameras, they do not naturally produce image frames for which the
overwhelming majority of today's image processing soft- and hardware are designed.

This paper describes an algorithm (FIBAR) for recovering a brightness image
from the events of an event camera. Each
individual event received from the camera updates the brightness image,
and as such, FIBAR is an asynchronous method compared to
synchronous methods that accrue multiple events and apply them all at
once\footnote{While FIBAR is in principle event-by-event, in practice it is not implemented on a
neuromorphic processor \cite{orchard2021loihi2}, meaning events arrive at the processing
host via USB bus in packetized form}.

Image reconstruction from events has been well studied before, but the vast majority of works
are not truly asynchronous because they first form a frame-type tensor that is then fed into
a neural network~\cite{scheerlink2020firenet}\cite{mostafavi2020}\cite{RebecqE2VID2019}\cite{eventhdr_2024}\cite{hypere2vid2024}
powered by a graphics processing unit (GPU).
This usually means committing to either a fixed time interval or a fixed number of events at the time
the tensor is constructed, thereby introducing a parameter that may have to be adjusted depending on the downstream task
or the scene.

In contrast, FIBAR is filter-based~\cite{pfrommer_2022_frequencycam}
and similar in spirit to~\cite{scheerlinck2019asynchronousconvolutions}.
It is particularly geared towards the situation where the camera is moving, but also works
satisfactorily~\cite{chaney_m3ed} when the camera is stationary.
FIBAR does not use time stamps directly and as such, is inherently timescale invariant.
No final update or filtering operations are required upon read-out, i.e.\ the reconstructed image
is always fresh.

FIBAR runs efficiently on a CPU and, as demonstrated here, yields sufficiently high-quality
reconstruction for detecting fiducial markers. The algorithm presented here could also be useful for other
tasks such as e.g.\ patch tracking in the context of sparse visual odometry, or could serve as a pre-processing
stage for learning-based approaches.

\section{Related work}
\label{sec:related_work}
Much work in the early literature leverages the DAVIS camera's ability to capture frame-based illumination intensities
as well as events, and focuses on reconstructing intermediate frames from the events, essentially performing an
interpolation in time. In \cite{brandli_2014} periodically arriving frame intensities are leveraged
to adjust event thresholds using an exponential moving average. This is similar to what FIBAR does, except that FIBAR does
not use frame-based intensities for reference, but rather assumes that in the long run, the intensity values at each pixel are stationary.
Ref.~\cite{fedi_2023} greatly improves the speed of image reconstruction, but still relies on the camera
to periodically provide intensity frames.

FIBAR draws heavily from earlier work on asynchronous spatial image convolutions\cite{scheerlinck2019asynchronousconvolutions}.
Both approaches share the idea of operating completely asynchronously, without employing any frames.
However, \cite{scheerlinck2019asynchronousconvolutions} uses an exponential time decay approach to
control noise (also in ~\cite{scheerlinckcontinuous2018}), which breaks timescale invariance, and implicitly assumes
that low event rates correspond to pixels with average illumination.
The latter is not a valid assumption.
FIBAR, on the other hand, presumes that pixels with low event rates correspond to areas with small image gradient.
In contrast to~\cite{scheerlinck2019asynchronousconvolutions}, where each event triggers an update for all pixels
within the aperture of a kernel, FIBAR's temporal filter and spatial filter only update the state at the
location of the event, thus
achieving high performance without a multithreaded implementation~\cite{rosa_2022}. 

The temporal digital filter deployed by FIBAR is described already in earlier work on Frequency Cam~\cite{pfrommer_2022_frequencycam},
where the zero-crossings of a per-pixel illumination intensity reconstruction are used to determine the frequency of a periodic signal.
For frequency estimation, the reconstruction filter is typically tuned towards smaller cutoff periods, which
allows faster frequency detection when the image is moving. Note that Frequency Cam does not perform any spatial filtering.

Ref.~\cite{manderscheid_2019_cvpr} served as inspiration for the spatial filtering algorithm. The authors describe
a way to maintain a speed invariant time surface, which, like FIBAR, does not make any direct use of time stamps, just
the order in which events occur. This leads directly to timescale invariance, but also results in persistent noise being impounded
into the time surface in areas where updates no longer occur. FIBAR addresses this issue with an event queue and spatial filtering.

Most recently, an event-based single integration (ESI) approach similar to FIBAR has been reported in
~\cite{dong2025eventbasedfastintensityreconstruction}, also presenting experiments on AprilTag detection.
Ref.~\cite{dong2025eventbasedfastintensityreconstruction} is particularly relevant because it quantitatively
compares their ESI algorithm to several other related schemes and GPU-based ones such as FireNet.
ESI differs from FIBAR in that it has no spatial filter, but rather accomplishes removal of noise with a time decay function, somewhat similar to ~\cite{scheerlinck2019asynchronousconvolutions}.
ESI essentially is a variant of the \texttt{libcaer}~\cite{libcaer} Accumulator module with a time decay rule that blends
linear and exponential decay.
It adapts the decay rate depending on the event rate, and makes explicit use of event time stamps, which FIBAR does not.

All learning-based image reconstruction methods begin by encoding time slices of events into dense tensors.
E2Vid~\cite{RebecqE2VID2019} and FireNet~\cite{scheerlink2020firenet} discretize the time axis within each tensor into five bins
and follow a voting scheme as proposed in ~\cite{alex_zhu_tensor_2019}. If the events follow each other closely enough in time,
such a voting scheme actually constitutes a summation of polarities (cf. Eq.~(\ref{eq:brightness_change})),
albeit one over a very short time horizon. The tensor is then fed into a recurrent encoder-decoder neural network that
is trained on synthetic data with a temporal consistency and VGG feature loss. The neural networks outperform~\cite{RebecqE2VID2019}
regularization-based methods\cite{munda_2018}, which also rely on GPU support. Later work~\cite{paredes_2021} showed competitive
reconstruction results when training in a self-supervised fashion, leveraging photometric consistency in combination with a network
for predicting optical flow. Noteworthy is also~\cite{wang_gan_2019}, which uses conditional
generative adversarial networks and discusses the effect of tensor construction from events.

The present paper makes no claim that its method is competitive with learning-based approaches in terms of reconstruction quality,
just that for some specific tasks, a much more lightweight approach will also achieve sufficiently high fidelity.
For this reason, only FireNet~\cite{stoffregen_2020} is used for a simple comparison, because its code and the network weights are readily available online.

For an in-depth discussion of calibrating event cameras from images reconstructed with E2Vid, see Ref.~\cite{muglikar_calib_2021}.
Methods that do not rely on image reconstruction but detect markers directly via feature extraction from events have reported
higher calibration accuracy\cite{salah_ecalib_2024}.
However, such methods rely on custom markers that are not supported by mainstream calibration toolboxes such as Kalibr\cite{kalibr_2013}.

\section{Notation}
The notation of this paper follows \cite{gallego_delbrueck_orchard} and \cite{pfrommer_2022_frequencycam}.
Here, the brightness $L$ of a given pixel is defined as the logarithm of its photo current $I$:
\begin{equation}
  L = \log(I)\ ,
  \label{eq:definition_brightness}
\end{equation}
and events $e_k = (t_k, p_k)$ with time index $k$ reported by the sensor indicate changes in brightness at time $t_k$
with polarity $p_k \in \{-1, +1 \}$ being positive if the brightness increased by more than a contrast threshold $\con$ and
negative if it decreased by more than a threshold $\coff$.

By defining the polarity-dependent contrast threshold
\begin{equation}
  \Cpki = \begin{cases}
    \Coffi &\text{if}\ p_k = -1\\
    \Coni  &\text{if}\ p_k = +1
  \end{cases}\ ,
  \label{eq:define_threshold}
\end{equation}
the brightness change due to event $k$ can be expressed as
\begin{equation}
  \deltaLki = \Cpki p_k
  \label{eq:brightness_change}\ .
\end{equation}

Note that due to fluctuations in the sensor's manufacturing process, the contrast thresholds depend on the pixel $i$,
see the discussion in Sec.~\ref{sec:pixel_thresholds}.

\section{Method}
\label{sec:method}
\subsection{Temporal digital filtering}
\label{sec:temporal_digital_filtering}
Much of the digital filtering algorithm described in this section has already been presented in \cite{pfrommer_2022_frequencycam} in the context of
frequency estimation. For a coherent reading experience, it will be reproduced here with slightly different notation.

An obvious way to estimate a pixel's brightness is to sum up the brightness changes observed up to time $t$,
taking into account the respective ON and OFF contrast thresholds $\Cpki$ at each pixel:
\begin{equation}
  \lcalib(t) = L_0 + \sum_{k | t_k \leq t} \Cpk p_k ,
\label{eq:C_reconstruction}
\end{equation}
where the pixel superscript $(i)$ is dropped to reduce notational clutter, and
the sum is understood to run only over those events that occur at the pixel under consideration.

The challenges with implementing (\ref{eq:C_reconstruction}) are:
\begin{enumerate}
  \item The contrast thresholds $\Cpk$ are not known a priori.
  \item The brightness $L_0$ that is present before the first event arrives is not known, and usually
not the same for all pixels. For a moving camera this leads to a persistent ghost image.
\item The camera has low-pass characteristics and a non-zero refractory time, meaning that events can get
lost during times of rapidly changing brightness.
\item Any noise events (thermal noise, shot noise) are compounded and retained indefinitely
in the reconstructed image.
\end{enumerate}

The first problem can be addressed by determining the contrast thresholds via
calibration \cite{brandli_2014}\cite{wang_ng_2019}. But this requires additional effort before deployment,
is not straightforward for cameras that have no frame readout (see the discussion in Sec.~\ref{sec:pixel_thresholds}),
and can be avoided as shown below.

The remaining problems 2.~-~4.\ are all handled here by temporal high-pass filtering of the signal, meaning that over time, the
initial ghost image and subsequently compounded noise will be "forgotten". Note that in the present context, {\em time} does
{\em not} refer to wall clock time (seconds), but to {\em event time}. Event time stamps are completely ignored, and instead,
each pixel operates on its own clock, which is advanced by one unit (or "sample" in the digital filtering literature)
when an event arrives at that pixel. This has the advantage that the resulting algorithm is inherently timescale invariant, i.e.\ 
it will operate the same way, irrespective of how fast the camera moves. This, of course, is just true to the extent
that the events emitted by the camera exhibit timescale invariance\footnote{Unfortunately, it is well known that due to various bandwidth limitations, the sensor produces fewer events as
the signal frequency increases \cite{lichtsteiner_posch_delbrueck_2008}\cite{shining_light_2023}.}. Operating on event time also implies that if a pixel
no longer gets events, its brightness value remains constant (except for potential filtering when pixel staleness is detected).

Assuming that the unknown initial brightness is $L_0 = 0$, and setting the unknown contrast thresholds $\Cpk = 1$ leads to the
simplest conceivable reconstruction method: adding up the polarities of the arriving events at each pixel independently:
\begin{equation}
  \lsimple_k = \sum_{k^\prime | k' \leq k} p_{k'} .
\label{eq:simple_reconstruction}
\end{equation}
Because updates occur only when events arrive, the continuous time argument $t$ from
Eq.~(\ref{eq:C_reconstruction})  has been dropped in Eq.~(\ref{eq:simple_reconstruction}) in favor of the
discrete event time index k.

\begin{figure}[t]
%
%
%
  \centering
  \includegraphics[width=1.0\linewidth]{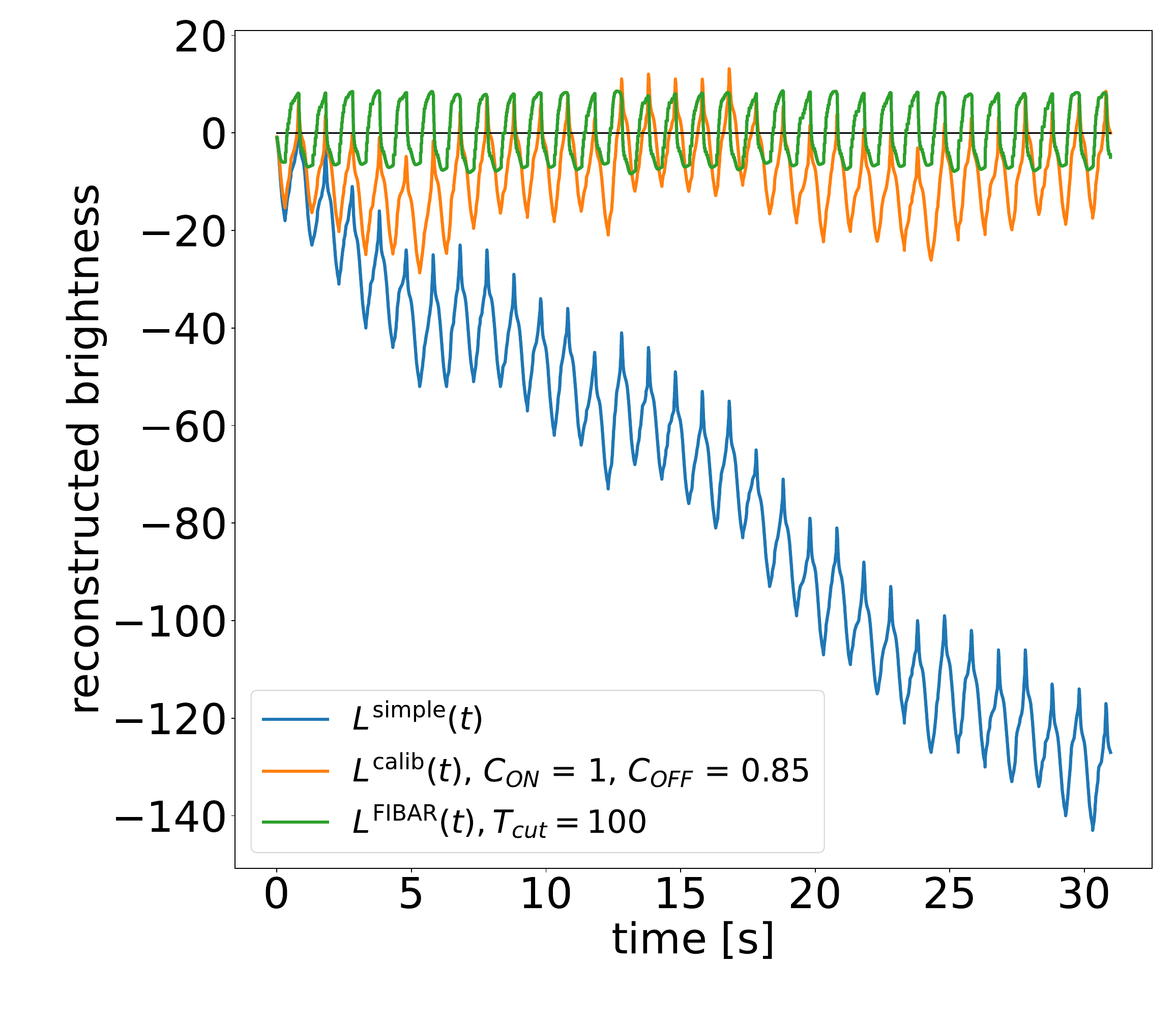}
  \caption{Reconstructed (scaled) brightness at the center pixel (319, 239) for a periodic signal (see Sec.~\ref{sec:pixel_thresholds}).
  $\lsimple(t)$ is obtained by a simple sum of the polarities, see Eq.~(\ref{eq:simple_reconstruction}).
  Unequal ON/OFF contrast thresholds result in more OFF than ON events and a downward trend of $\lsimple(t)$.
  $\lcalib(t)$ has been detrended by adjusting the threshold $\coff$, but still,
  low-frequency oscillations are present.
  FIBAR filtering with a cutoff period of $\Tcut=100$ removes both the trend and the low frequency distortions.}
  \label{fig:periodic_signal}
\end{figure}

Fig.~\ref{fig:periodic_signal} shows both the simple and the contrast threshold adjusted reconstruction
for a single pixel of the dataset described in Sec.~\ref{sec:pixel_thresholds}. Correcting $\lsimple$ by accounting for
the ON/OFF threshold imbalance removes the trend ($\lcalib$ in Fig.~\ref{fig:periodic_signal}), but this requires pre-processing
the entire dataset.

Removing trend and low-frequency oscillations is accomplished by the FIBAR algorithm as follows.
First, any imbalance between ON and OFF events is compensated for
by subtracting from $\pk$ its moving average $\pbark$. Upon arrival of event $k$, rather than adding $p_k$ as
in Eq.~(\ref{eq:simple_reconstruction}), one adds the detrended brightness increment
\begin{equation}
  \label{eq:deltaL_det}
  \deltaLdet_k = \pk - \pbark
\end{equation}
to the sum. Here, $\pbark$ is the exponential moving average\cite{pfrommer_2022_frequencycam}
\begin{equation}
  \label{eq:exp_average}
  \pbark = \alpha\ \bar{p}_{k-1} + (1 - \alpha)\ \pk
\end{equation}
with a mixing coefficient $\alpha \in [0, 1]$.
This mixing coefficient will need to be set sufficiently large to capture enough ON and OFF events
for estimating the ON/OFF threshold imbalance, but not so large as to delay the estimation
unnecessarily. The proper choice of $\alpha$ is discussed below once the full filter algorithm has been presented.

Using the detrended brightness increment leads to a modified Eq.~(\ref{eq:simple_reconstruction}) for the detrended
reconstructed brightness, expressed recursively:
\begin{equation}
  \label{eq:detrended_brightness}
  \Ldet_k = \Ldet_{k-1} + \deltaLdet_k
\end{equation}

After detrending, $\Ldet$ is further filtered with a high-pass filter:
\begin{align}
\Lfibar_k &=\beta \Lfibar_{k-1} + \frac{1}{2}(1 + \beta)(\Ldet_k - \Ldet_{k-1}) \label{eq:high_pass_1}\\
& = \beta \Lfibar_{k-1} + \frac{1}{2}(1 + \beta)\deltaLdet_k \label{eq:high_pass_2}\ .
\end{align}
The differencing in the second term of (\ref{eq:high_pass_1}) immediately undoes the integration
in (\ref{eq:detrended_brightness}) and leads to Eq.~(\ref{eq:high_pass_2}).
The particular way the filter parameter $\beta \in [0,1]$ is introduced in (\ref{eq:high_pass_1})
asserts unit gain at maximum frequency ($\omega = \pi$).
Combining equations (\ref{eq:deltaL_det}), (\ref{eq:exp_average}), and (\ref{eq:high_pass_2})
by means of the $z$-transform\cite{proakis_manolakis_2006}
\begin{equation}
Y(z) = \sum_{n=-\infty}^\infty y_nz^{-n} \label{eq:z_transform}
\end{equation}
yields the relationship between reconstructed brightness $\Lfibar(z)$ (output)
and event polarity $P(z)$ (input):
\begin{align}
  \Lfibar(z)&= H(z) P(z) = H_\alpha(z) H_\beta(z) P(z)\label{eq:transfer_functions}\\
  H_\alpha(z)&= \frac{\alpha(z-1)}{z-\alpha}\\
  H_\beta(z)&=\frac{z(1+\beta)}{2(z-\beta)}\ .
 \end{align}
 For implementation, this could be written as a single recursive filter:
 \small
 \begin{equation}
  \Lfibar_k = (\alpha + \beta)\Lfibar_{k-1} - \alpha\beta\Lfibar_{k-2} +\frac{\alpha}{2}(1+\beta)(p_k - p_{k-1})\ ,
  \label{eq:fibar_iir}
 \end{equation}
\normalsize
but in practice it is more convenient to update $\pbark$ according to (\ref{eq:exp_average}),
then compute $\deltaLdet_k$ (\ref{eq:deltaL_det}) and update $\Lfibar_k$ following (\ref{eq:high_pass_2}).
The latter is preferred because, for spatial filtering (Sec.~\ref{sec:spatial_filtering}),
only $\Lfibar_k$ is filtered, but $\pbark$ is not.

\begin{figure}[t]
  \centering
  \includegraphics[width=1.0\linewidth]{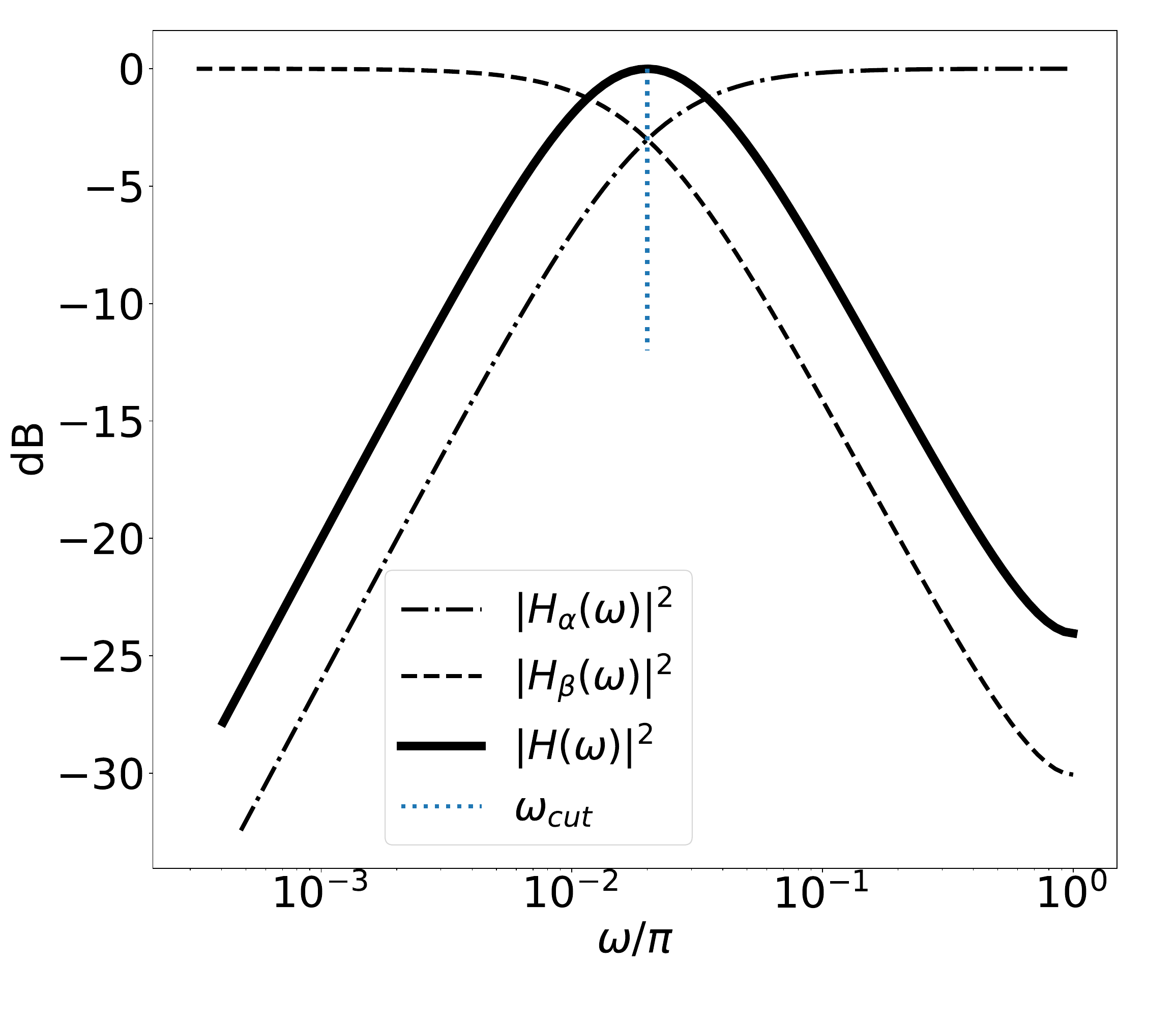}
  \caption{Bode magnitude plot of the transfer functions in Eq.~(\ref{eq:transfer_functions}).
  The cutoff frequency of $H_\alpha$ and $H_\beta$ has been set to coincide at $\ocut = 0.02\pi$
  which corresponds to $\Tcut=100$ events. The band pass center frequency is also close to $\ocut$.}
  \label{fig:transfer_functions}
\end{figure}

The transfer function $H(z)$ is the product of
a high-pass $H_\alpha(z)$ and a low-pass\footnote{Although the averaging of (\ref{eq:exp_average}) constitutes a low pass,
the result is subsequently subtracted from an all pass (\ref{eq:deltaL_det}), rendering $H_\alpha$ a high pass.
Similarly, the integration (\ref{eq:detrended_brightness}) is a low-pass operation that compensates for the high-pass
filtering in Eq.~(\ref{eq:high_pass_1}) to make $H_\beta$ a low pass.} $H_\beta(z)$,
resulting in a band pass as shown in Fig.~\ref{fig:transfer_functions}.

How should the filter coefficients $\alpha$ and $\beta$ be set? To answer this question, it is important to first get
a better understanding of the signal, which is actually a sequence of polarities $p_k$ occurring in event time. Rather
than considering the frequency $\omega$, it is more intuitive to think in terms of the period $T= 2\pi/\omega$,
which is expressed in number of events.

In any image reconstruction, it is implicitly assumed that there will be only a finite number of ON or OFF events before
the sensor's or the scene's dynamic range is reached. Afterwards, events of opposite polarity can be expected,
totaling up to $\nrange = \non + \noff$ events for a single sweep over the dynamic range. Clearly, the cutoff
period $\Tcut = 2\pi/\ocut$ must be longer than $\nrange$, such that the ON/OFF threshold imbalance can be determined.
In practice, it is found that
\begin{equation}
\Tcut = 4 \max(\noff, \non) \label{eq:tcut}
\end{equation}
is a good choice. For the sensors deployed in this work, $\Tcut \geq 40$ works well for scenes with natural light, but
when the full dynamic range of the sensor is used, such as in Fig.~\ref{fig:periodic_signal},
a larger $\Tcut=100$ yields better results because a premature saturation of the reconstructed brightness is avoided.
For a periodic signal like in Fig.~\ref{fig:periodic_signal} it is straightforward
to determine $\non$ and $\noff$ by dividing the number of ON and OFF events by the number of cycles.
For additional experiments on how $\Tcut$ affects the reconstruction, see \cite{pfrommer_2022_frequencycam}.

In the first stage of the filter, $\alpha$ is chosen such that $H_\alpha$ passes a signal with period $\Tcut$,
i.e.\ $|H_\alpha(\exp(j\omega))|^2$ assumes half of its maximum value at $\ocut$:
\begin{equation}
  \alphacut = \frac{1-\sin{\ocut}}{\cos{\ocut}}\ . \label{eq:alphacut}
\end{equation}
The coefficient $\beta$ must now be chosen such that the second stage of the filter passes the signal from the first stage,
but filters any higher frequency noise.
Hence, $\beta$ is selected to set the cutoff frequency of $H_\beta(\exp{(j\omega)})$ to $\ocut$ as well, implying:
\begin{equation}
  \betacut = (2 - \cos{\ocut})- \sqrt{(2-\cos{\ocut})^2 - 1}\ . \label{eq:betacut}
\end{equation}
This completes the recipe for picking $\alpha$ and $\beta$ depending on the anticipated signal property $\Tcut$.
Usually $\ocut \ll 1$ holds well, and since both $\alphacut$ and $\betacut$ have identical Taylor series expansions of
$1 - \ocut + \frac{1}{2}\ocut^2 + \mathcal{O}(\ocut^3)$ to second order, one finds $\alphacut \approx \betacut$ 
and $H(z)$ having a double pole on the real axis at $z = \alphacut$.

To summarize the implementation: pick a $\Tcut$ that is suitable for the signal considered (\ref{eq:tcut}), compute $\ocut = 2\pi / \Tcut$,
then $\alphacut$ (\ref{eq:alphacut}), $\betacut$ (\ref{eq:betacut}), and finally obtain the reconstructed brightness with the IIR
filter in Eq.~(\ref{eq:fibar_iir}).

\subsection{Spatial filtering}
\label{sec:spatial_filtering}
In addition to the temporal filtering described in Sec.~\ref{sec:temporal_digital_filtering},
FIBAR also implements a spatial filter to reduce the noise in the reconstructed image.
For a moving camera, pixels that are not producing events should have comparatively
small image gradients, and hence FIBAR applies a one-time 3x3 Gaussian spatial filter to
a pixel that stops producing events (the pixel becomes "stale").

How to detect stale pixels? Using a per-pixel timer would mean committing to a timescale and therefore
abandoning timescale invariance. Filtering pixels after a fixed number of events have occurred
will avoid committing to a timescale, but the number of events to wait before filtering a pixel
depends on the texture of the scene. If an event has just occurred at a given pixel, 
a richly textured scene will generate events in many other locations before that pixel can be
considered stale, whereas a low texture scene (e.g.\ with a single image feature) would call for
marking the pixel stale after just a few events.

FIBAR addresses this by maintaining a global LIFO queue with events for all active pixels ("active event queue"),
and by performing spatial filtering on a pixel at the time its last event is removed from the queue.
The target length $\qtarg$ of the queue is dynamically regulated to account for the texture of the scene.

Postponing the discussion of how $\qtarg$ is regulated and assuming for the moment $\qtarg$ as given,
the spatial filtering works as follows when an event arrives for a pixel:
\begin{enumerate}
  \item The brightness at the pixel location is updated employing the temporal filter
  from Sec.~\ref{sec:temporal_digital_filtering}.
  \item The event is added to the back of the event queue, and a per-pixel counter is
  incremented to keep track of how many events for that pixel are in the queue.
  \item Old events are removed from the front of the queue until the queue
   has reached $\qtarg$. The per-pixel event counter is decremented accordingly, and
   if a pixel's event counter reaches zero, it becomes inactive and is spatially filtered.
\end{enumerate}
The per-pixel active event counter is necessary to handle the frequent scenario where 
multiple events occur at a pixel in rapid succession.

\begin{figure}[t]
  \centering
  \includegraphics[width=1.0\linewidth]{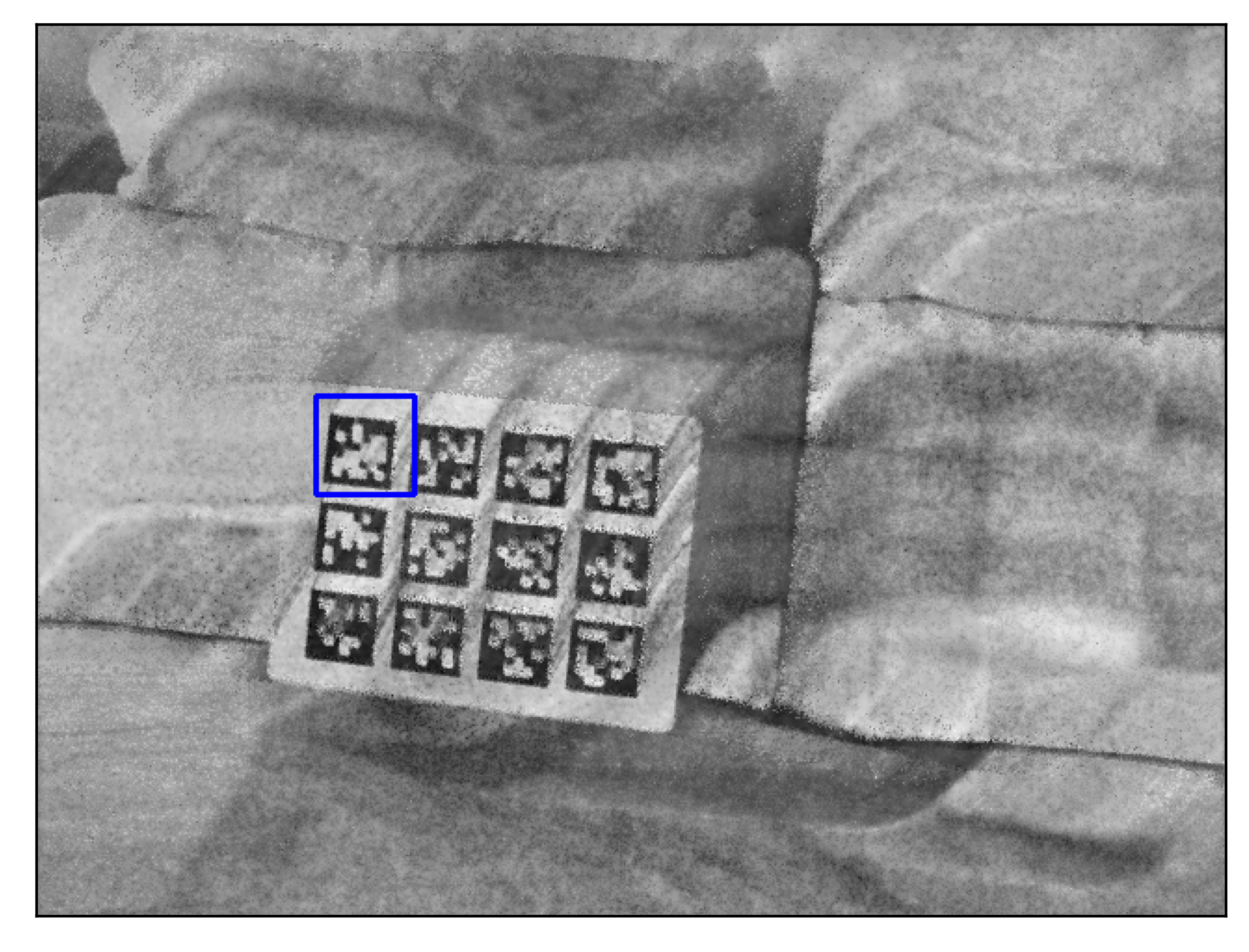}
  \caption{Frame reconstructed with $\rfill=0.5$ and $\Tcut=40$. The marked blue rectangle is
  shown in Fig.~\ref{fig:fill_ratios}.}
  \label{fig:fill_ratios_frame}
\end{figure}

\begin{figure}[t]
  \centering
  \includegraphics[width=1.0\linewidth]{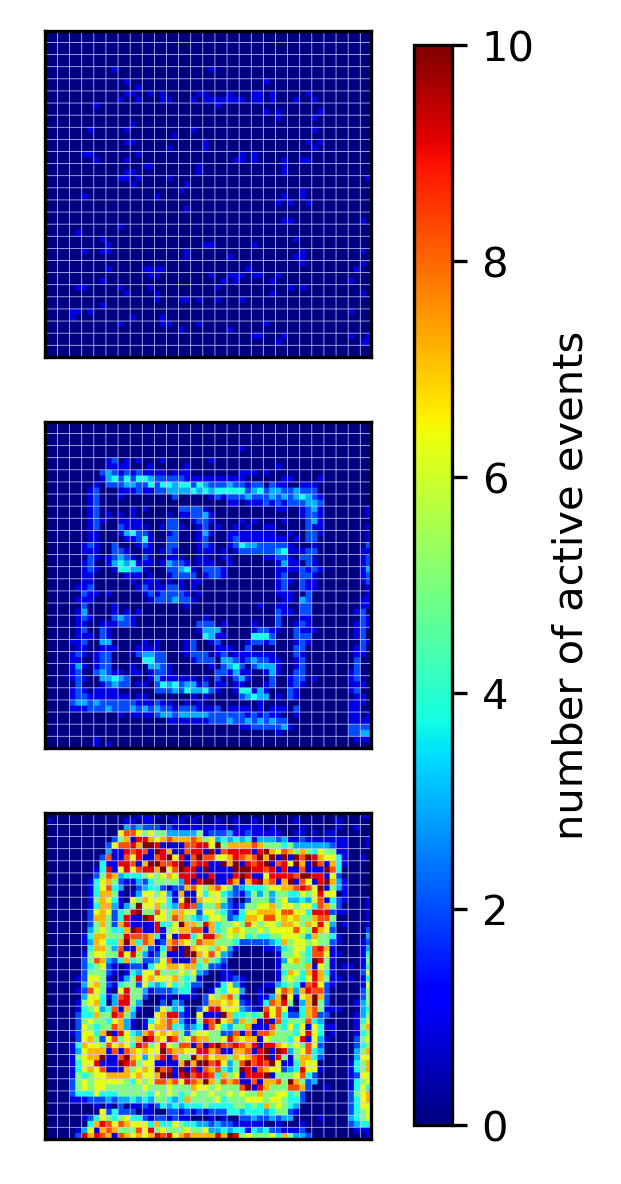}
  \caption{Image of active pixels (IAP) for fill ratios 0.3 (top), 0.5 (middle) and 0.7 (bottom) for
  the rectangle marked in Fig.~\ref{fig:fill_ratios_frame}.
  Color indicates the number of active events at each pixel.
  The length of the active event queue grows with the fill ratio, resulting in multiple active
  events per pixel (middle and bottom). Too low a fill ratio (top) leads to premature spatial filtering of pixels that
  are still near strong image gradients and will generate further events in the near future.
  If the fill ratio is too high (bottom), the pixel remains marked as active for so long
  that close-by image features are already generating events again, thereby delaying the
  spatial filtering unnecessarily.}
  \label{fig:fill_ratios}
\end{figure}

The remaining question is how to set the target length $\qtarg$ of the active event queue.
Assuming the case of a camera moving in a scene dominated by edge
features\footnote{This is clearly not a good assumption in outdoor natural scenes that display
grass or trees with foliage. In this case, the queue of active events will grow very long,
and spatial filtering will be rare. Whether this is detrimental to reconstruction requires
further study.} a reasonable criterion is that the image formed by all active pixels (IAP) not
be blurry. The IAP is defined as an image where each pixel's value equals the number of
active events for that pixel.
As the number of active pixels increases, the AIP becomes increasingly blurry,
see Figs.~\ref{fig:fill_ratios_frame} and \ref{fig:fill_ratios}. Hence, the length of $\qtarg$ is regulated
such that the IAP remains sharp.

%
%

To measure the sharpness of the AIP, it is tiled with a 2x2 pattern, see Fig.~\ref{fig:fill_ratios},
and the fill ratio
\begin{equation}
  \rfill = \frac{\Nap}{\Nat A} \label{eq:fill_ratio}
\end{equation}
is computed, where $\Nap$ is the total number of active pixels, $\Nat$ is the number of active tiles, i.e.\ the
tiles that have at least one active pixel, and $A$ is the area of the tile, in this case $A=2\times 2=4$.
Note that a larger fill ratio implies a blurry image, but the fill ratio is at least $1/A = 0.25$
and at most $1$. Increasing the length of the active event queue $\qtarg$ increases the
number of active pixels, and therefore the fill ratio and the blurriness of the AIP.

The update of $\qtarg$ occurs right after all old events have been processed
and removed from the queue. If the current step is indexed with $k$, and the current
target queue length and the currently observed fill ratio are called $\qtarg_k$ and $\rfill_k$,
respectively, then the target queue length is updated based on the current queue length $q_k$ as follows:
\begin{equation}
  \label{eq:queue_length_update}
  \qtarg_{k+1} = \lfloor q_{k}\frac{\rfill}{\rfill_k}\rfloor\ .
\end{equation}
If $\rfill > \rfill_k$, the target queue length is increased, which will increase the amount of
blur and therefore the observed fill ratio. This way, the observed fill ratio $\rfill_k$
is kept close to the target $\rfill$. In practice, (\ref{eq:queue_length_update})
is implemented using $\Nap$ and $\Nat$ directly to avoid floating point arithmetic. Note that because of the
floor operation in (\ref{eq:queue_length_update}), $q_k$ must be suitably bounded from below to
avoid the queue collapsing to small
values.\footnote{Consider the extreme case of $q_k=0$, where the queue length will evidently not be able to grow again}

By experimenting with $\rfill$ (see Fig.~\ref{fig:fill_ratios}) it is found that $\rfill=0.5$ works well in practice
and strikes a good compromise between spatially filtering pixels too early or too late.

\section{Experiments}
\label{sec:experiments}
\subsection{Tag detection}
\label{sec:tag_detection}
This section presents experiments demonstrating that FIBAR can reconstruct image frames from events
sufficiently well to detect fiducial optical markers.
To this end, a dataset is recorded with a SilkyEVCam camera employing a Computar 8~mm lens
and a Prophesee Gen 3.1 sensor with
640x480 pixels resolution. All biases are left at their defaults
as configured by the OpenEB 5.0 SDK. The fiducial markers are 70~mm wide AprilTags of the family 36h11\cite{wang_apriltags_2016},
printed in a three-by-four grid on a flat 1/8~in thick dibond aluminum board with a tag spacing of 17~mm.
The camera is moved around in the room while keeping the board in view and recording the event stream using the
open source ROS2 Metavision driver and the ROS2 bag recorder running together as a
composable node to prevent data loss. Subsequently, image frames are reconstructed from the recorded event stream at 40 fps using the FIBAR
algorithm described in Sec.~\ref{sec:method}. The dataset consists of about 296 million events that are reconstructed into 1314 frames.
The UMich AprilTag library version 3.4.2\cite{wang_apriltags_2016} was used to detect tags in the reconstructed images.

The first step is to explore how the cutoff period $\Tcut$ from Eq.~(\ref{eq:tcut}) and the
fill ratio $\rfill$ from Eq.~(\ref{eq:fill_ratio}) affect the reconstruction (cf.\ Tab.~\ref{tab:parameter_scan}).
It is found that indeed the spatial filtering improves the detection rate significantly, and that values of $\rfill = 0.5$
and $\Tcut = 40$ work well. These parameter choices were then used for all subsequent experiments.

\begin{figure}[t]
  \centering
  \includegraphics[width=1.0\linewidth]{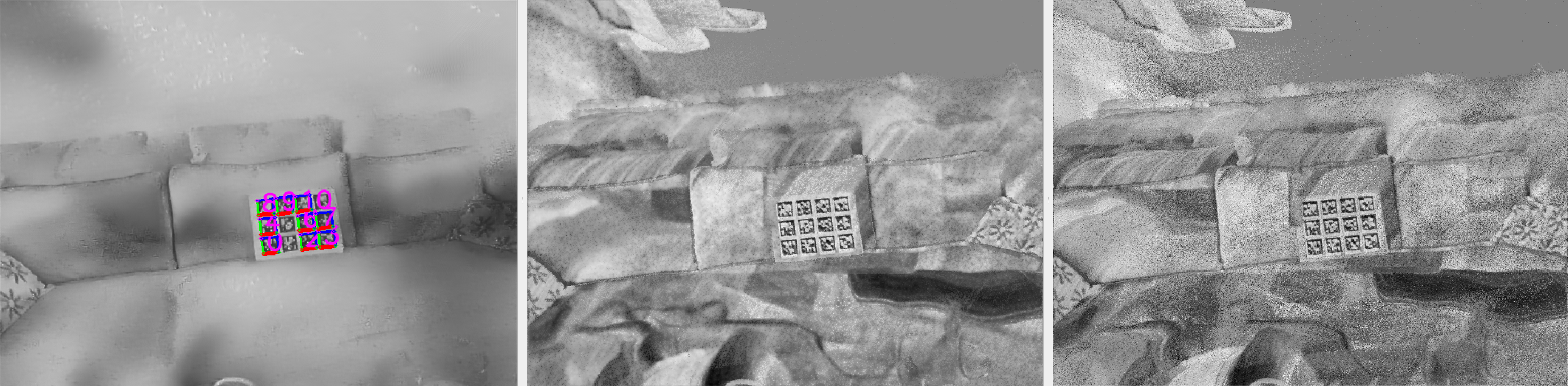}
  \includegraphics[width=1.0\linewidth]{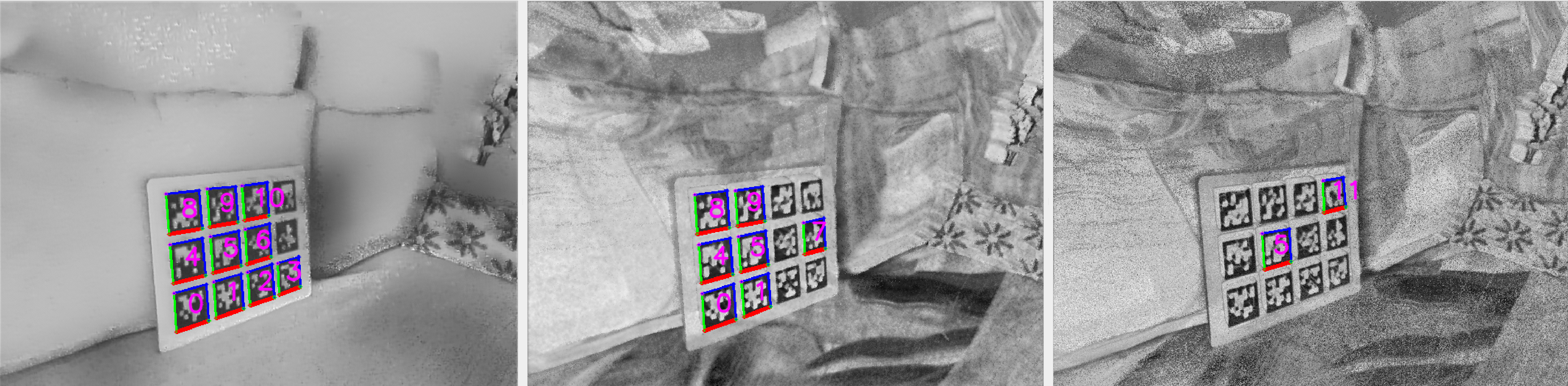}
  \includegraphics[width=1.0\linewidth]{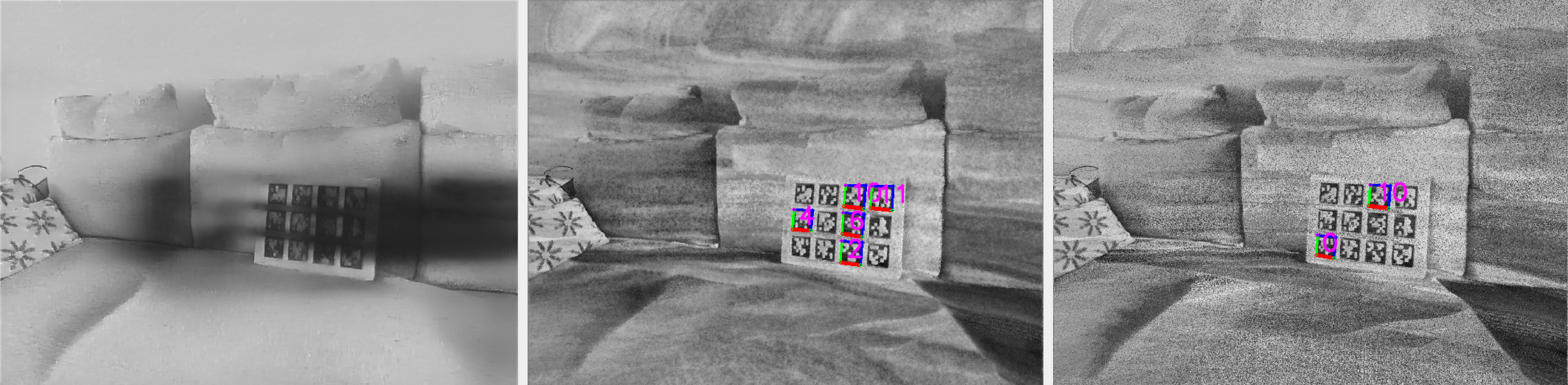}
  \caption{Frames reconstructed with FireNet (left column), FIBAR (center column),
           and FIBAR without spatial filtering (right column).
           FireNet reconstructions exhibit much lower noise and therefore allow the
           detection of small tags further away, but occasionally suffer from
           persistent artifacts that prevent tag detection. Compare the
           center and right column to see the value of spatial filtering.}
  \label{fig:tag_detection_frames}
\end{figure}

With the correct parameter settings determined, a comparison with FireNet can now be made, see Fig.~\ref{fig:tag_detection_frames} and
Tab.~\ref{tab:tag_detection}. For the dataset taken here, FireNet actually detects fewer tags than FIBAR.
While the FireNet reconstructions have much less noise and allow for the detection of smaller tags, they are
at times marred by dark artifacts that inhibit tag detection (bottom left corner of Fig.~\ref{fig:tag_detection_frames}).
Being only 32 seconds long, the dataset presented here is too small and lacks the diversity necessary for a quantitative assessment
of one method vs.\ the other. However, it does demonstrate that for this specific task, FIBAR is at least competitive with established
methods.

\begin{table}
  \centering
  \begin{tabular}{c|c|c}
    $\Tcut$ & $\rfill$ & \# tags detected\\
    \hline
    5&0.5&0\\
    10&0.5&18\\
    20&0.5&569\\
    30&0.5&1373\\
    40&0.5&1705\\
    50&0.5&1715\\
    60&0.5&1658\\
    100&0.5&1396\\
    40 & no filter&834\\
    40 & 0.3 & 1472\\
    40 & 0.7 & 1073\\
  \end{tabular}
\caption{Number of AprilTags detected for different FIBAR parameter settings. A smaller value for
$\Tcut$ leads to a faster estimation of the threshold parameters, but too small a $\Tcut$ results
in premature pixel saturation.
The remainder of this work uses settings of $\Tcut = 40$ and $\rfill = 0.5$.}
  \label{tab:parameter_scan}
\end{table}

\begin{table}
  \begin{tabular}{c|c|c|c}
        &FireNet & FIBAR & FIBAR-NSF\\
    \hline
    \# tags detected  & 1215 & 1705& 834\\
  \end{tabular}
\caption{AprilTag detection statistics for different reconstruction methods. FIBAR-NSF means FIBAR without
spatial filtering. The dataset consists of about 296 million events that are reconstructed into 1314 frames at 40fps.
FIBAR detects more tags because FireNet reconstruction artifacts can impede tag
detection (see Fig.~\ref{fig:tag_detection_frames}).}
  \label{tab:tag_detection}
\end{table}

\subsection{Inter-pixel threshold variation}
\label{sec:inter_pixel_variation}
As outlined in Sec.~\ref{sec:temporal_digital_filtering}, FIBAR compensates for a pixel having different thresholds $\Coni$ and $\Coffi$,
but {\em does not} account for threshold variations {\em between} pixels. Could the image reconstruction be materially improved
by scaling the reconstructed brightness of each pixel by its threshold $\Cip$ (Sec.~\ref{sec:pixel_thresholds}, Eq.~\ref{eq:cip})
relative to other pixels?
To answer this question, the relative pixel thresholds $\Cip$ were determined by exposing the sensor to a spatially uniform,
periodic illumination and subsequently counting the number of ON and OFF events for each pixel.
Correcting the reconstruction by multiplying the r.h.s.~of Eq.~(\ref{eq:deltaL_det}) with $\Cip$ yielded {\em no material improvement} in the
tag detection rate. See Sec.~\ref{sec:pixel_thresholds} for more details and a discussion of this somewhat surprising result.

\subsection{Calibration}
\label{sec:calibration}
\begin{figure}[ht]
  \centering
  \includegraphics[width=0.8\linewidth]{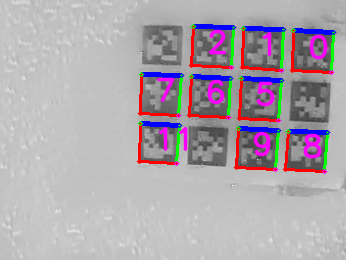}
  \includegraphics[width=0.8\linewidth]{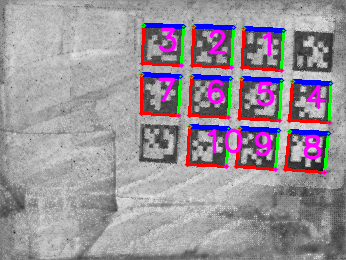}
  \includegraphics[width=0.8\linewidth]{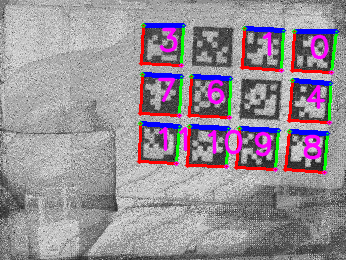}
  \caption{Example calibration frames, reconstructed from DAVIS 346 events using FireNet (top), FIBAR (middle),
  and FIBAR without spatial filtering (bottom).
  Note the error in the location of the detected tag corners for e.g.\ tag 1
  in the middle image and tag 0 in the bottom image.}
  \label{fig:calibration_frames}
\end{figure}

The experiments in this section show that FIBAR is suitable for camera calibration and produces results that are
very similar to the ones obtained by FireNet.
In contrast to the previous section, the dataset used here is collected with an Inivation DAVIS 346 camera
at default bias settings with a resolution of 346x260 pixels.
The AprilTag board and the tag detection library used are the same as described in Sec.~\ref{sec:tag_detection},
however, the camera remains stationary, and the board is moved instead.
The dataset spans over 147 seconds, but owing to the lower resolution of the DAVIS camera and
different threshold bias settings, contains only 4.8 million events.
The event stream is reconstructed into 5872 frames corresponding to a rate of 40~Hz.
As a cross-check, a second dataset is taken with the Davis camera's APS feature (frame-based sensor readout) enabled\footnote{The read-out of the APS frames introduces
additional noise into the event stream, which is why this feature was disabled when recording events.}.
This dataset spans 200 seconds and contains 1016 frames.
Using the UMich AprilTag detector library again to extract the corners of the tags from the reconstructed (or APS)
frames, a calibration is performed with the ROS2-based \texttt{multicam\_imu\_calib} package, which minimizes the reprojection
error under a Huber norm with parameter 1.345. The results are presented in Tab.~\ref{tab:calibration_results},
with reconstructed sample images shown in Fig.~\ref{fig:calibration_frames}.

Similar to the results in Sec.~\ref{sec:tag_detection}, FIBAR (with spatial smoothing) detects more corners than FireNet.
As expected, the APS frames have the lowest reprojection error (second row in Tab.~\ref{tab:calibration_results}),
followed by FireNet and FIBAR (with and without spatial smoothing). Fig.~\ref{fig:calibration_frames} illustrates
the source of such noise. However, despite the added noise, the calibration parameters
ultimately obtained are very similar for all three event reconstruction methods (FireNet, FIBAR, FIBAR-NSF).
For focal lengths, the numbers are off by less than 1\% from the APS frame results, and for the image center, the
largest error is a 2.8\% deviation in $c_x$ for FIBAR-NSF.
Distortion coefficients are generally more difficult to estimate, and consequently, the discrepancies from the APS
results are larger. For instance, $k_1$ obtained by FIBAR is off by 9.7\%, and the FireNet $k_2$ is off by 16.2\%.
Comparable deviations have been reported in Ref.~\cite{muglikar_calib_2021}.

%
%
%

\begin{table}
  \begin{tabular}{l|S[table-format=3.3]|S[table-format=3.3]|S[table-format=3.3]|S[table-format=3.3]}
        &{APS}&{FireNet} &{FIBAR} & {FIBAR-NSF} \\
        \hline
    \#tags& {9031} & {43763} & {46052} &{39531}\\
    error& 0.3 & 0.58 &0.77 &0.73\\
    $f_x$ & 430.4 & 427.2 &427.7 &427.6\\
    $f_y$ & 429.7 & 425.6 & 425.9 & 425.9\\
    $c_x$ & 163.5 & 167.4 & 167.5 & 168.0\\
    $c_y$ & 126.3 & 124.1 & 124.8 & 125.0\\
    $k_1$ & -0.246 & -0.254 & -0.270 & -0.266\\
    $k_2$ &  0.212 &  0.177 &  0.215 &  0.205\\
    $p_1$ & -0.001 &  0.000 & -0.000 & -0.001\\
    $p_2$ &  0.001 &  0.001 &  0.001 &  0.001\\
  \end{tabular}
\caption{Calibration results obtained using the DAVIS 346 camera's APS frames and different reconstruction methods.
FIBAR-NSF means FIBAR without spatial filtering. The second row shows the root-mean-square reprojection error in pixels.
The radial-tangential calibration parameters are named following the OpenCV convention.}
\label{tab:calibration_results}
\end{table}

\subsection{Runtime performance}
\label{sec:runtime_performance}
To assess the speed of the proposed algorithms, the dataset for tag detection described in Sec.~\ref{sec:tag_detection}
is used. It consists of 296 million events from a 640x480 resolution sensor that are reconstructed into 1314 frames. 

For modern CPUs, performance is often determined by the memory access pattern and whether data needs to be fetched
from main memory or can be retrieved from cache. {\em For FIBAR to perform well, it is crucial that the
reconstruction state fit completely into cache memory}. Maintaining the filter state requires two
floating-point (4-byte) variables: $\pbark$ from Eq.~(\ref{eq:exp_average}) and $\Lfibar_k$ from Eq.~(\ref{eq:high_pass_2}).
The spatial filtering requires keeping track of the number of active events in the queue,
adding another 2-byte unsigned integer to the state. The compiler pads this structure from 10 bytes to 12
for alignment purposes, so the one-byte per-tile unsigned integer used to keep track of
the number of active pixels in a tile is stored in the filter state of the pixel at the
top corner of the tile without consuming extra memory.
By somewhat arbitrarily limiting the length of the queue of active events to the number of pixels,
the queue memory stays below 4 bytes per pixel, since a queue entry only takes up 2 bytes for x and y coordinates each.

All performance experiments presented here are single-threaded. While exploiting parallelism should be possible, parallelizing
the decoding of the compressed events is not straightforward due to the incremental way in which event time stamps are maintained
in the EVT3 compressed format. Parallelizing the state update is also non-trivial because data access latencies dominate over
compute time, meaning that cache invalidation due to data writes from multiple cores can easily negate any performance gains.

For the experiments presented here, an HP OmniBook Ultra with 32~GB of RAM and an AMD Ryzen AI9HX375 CPU is used.
It has 12 cores with a base clock of 2~GHz and a maximum boost clock of 5~GHz. The L1 cache with a
per-core size of 567~KB (data) and 384~KB (instructions) is followed by a per-core 12~MB unified L2 cache, and finally a combined
unified 24~MB L3 cache. Experiments are run on one of the four Zen 5 performance cores rather than the slower Zen 5c
efficiency cores, which reach only about 65\% of the speed of the performance cores. With a padded per-pixel state of 12 bytes,
and at most 4 bytes for the event queue, the entire state of FIBAR consumes at most 4.7~MB for the VGA sensor and thus
fits comfortably into the 12~MB L2 cache.

Tests on an Intel desktop CPU (i7-14700K at 5.6GHz) showed unimpressive performance gains of 12\% over the AMD laptop processor,
demonstrating the vanishing gap between mobile and desktop processors for single-threaded workloads.

\begin{table}[ht]
\setlength{\tabcolsep}{0.1cm}
\begin{center}
\begin{tabular}{P{1.2cm}|P{1.0cm}|P{1.3cm}|P{1.2cm}|P{1.3cm}|P{1.0cm}|}
                       &decode&temporal filter &active pixel tracking &Gaussian blur&total\\
                       \hline
        time [ns/ev]&4.5& 2.7&13.8 & 3.0&24.0\\
        \hline
        rate [Mev/s]&223 & 376 & 72 & 330& 42\\
  \end{tabular}
\end{center}
  \caption{AMD laptop Ryzen AI9HX375 CPU time consumption per event and implied rate (in million events/s)
  for EVT3 decoding, temporal filter update (Eq.~(\ref{eq:exp_average}) and (\ref{eq:high_pass_2})),
  active pixel tracking for spatial filtering (Sec.~\ref{sec:spatial_filtering}),
  and applying Gaussian blur for spatial filtering. Without event decoding, the time is 19.5~ns/ev (51~Mev/s).
  With decoding, but without spatial filtering (FIBAR-NSF) the reconstruction takes only 7.1~ns/ev (140~Mev/s).}
  \label{tab:performance}
\end{table}

As shown in Tab.~\ref{tab:performance}, the tracking of the active pixels is the most time-consuming step, taking about 14~ns/ev.
The temporal filter itself is very fast with only 2.7~ns/ev. Also expensive is the decoding of the events that are presented
in EVT3 format, although this code path has room for further optimization. End-to-end, the FIBAR algorithm runs at
 a rate of 42~Mev/s with spatial filtering, and at 140~Mev/s without.

For comparison, FireNet processes the entire dataset within 33.45~s on an
Intel i7-14700K CPU with an NVIDIA GeForce GTX~1070~Ti, implying a time of 112~ns/ev (8.8~Mev/s).
A direct comparison of the two approaches makes little sense, because in contrast to FIBAR,
FireNet is frame-based and thus the effort scales with the frequency of readouts. Each readout requires an
inference step that takes 15.4~ms/frame, such that at 40~fps, the inference step already takes up around 60\%
of the time.

\section{Limitations}
\label{sec:issues}
As much as timescale ("speed") invariance is a good and valid concept, it has its limits, because the
pixel's response has low-pass characteristics \cite{shining_light_2023} and therefore is {\em only approximately}
timescale invariant for relatively slow camera motions.
Experimentally, it is found that the ratio of the number of events to the magnitude of optical flow decreases with speed.
For this reason, image reconstruction algorithms must take camera speed into account in some form or another.

Another point of critique of FIBAR is the use of a single queue for tracking the active events. From an aesthetic point
it does not fit in well with neuromorphic concepts such as sparse and distributed computing. A central queue is also hard
to parallelize.

Further, the single event queue employed implicitly assumes an approximately uniform event rate across
the entire sensor. If, for example, two point features were to move at very different speeds, the slower feature's
events would be taken off the event queue too soon, and the Gaussian blurring would occur prematurely.

The preceding two limitations could be addressed by tiling the image and operating event queues separately for each tile.
However, one then has to think about how to advance the queues of tiles that no longer register any events.

In its current form, FIBAR does not take into account the magnitude of the gradient of pixels when performing spatial filtering.
One could consider enqueuing events again if there is still a large gradient at the pixel location after spatial filtering,
thus triggering another Gaussian blur at a later time.
Running a spatial filter on pixels that are immediate neighbors could also be beneficial. Unfortunately, any additional
operations, in particular when they involve updating the event queue, will negatively affect the runtime performance.

Another valid point of critique is that the spatial blurring does not take into account the {\em direction} of
the gradient. The absence of events at a pixel while the camera is moving only implies a small image gradient component
{\em parallel} to the direction of the optical flow. Gaussian blurring, however, is isotropic and will reduce image gradients
also in the direction {\em perpendicular} to the optical flow.

In view of its many conceptual deficiencies, the effectiveness of the spatial filtering (Tab.~\ref{tab:tag_detection})
is surprising. The adoption of more sophisticated algorithms will hinge on the ability to harness more powerful
compute hardware, such as GPUs or FPGAs.

\section{Conclusion}
\label{sec:conclusion}
This paper presents FIBAR, an algorithm to asynchronously reconstruct intensity images from event streams.
FIBAR first integrates the event polarities separately for each pixel using a digital filter.
Subsequently, pixels in the reconstructed image that have not emitted events for a while are spatially
filtered by Gaussian blurring. The resulting algorithm runs efficiently on a laptop CPU at 42 Mev/s and produces
images that can be used for downstream tasks such as the detection of fiducial markers.

\section{Appendix on relative contrast thresholds}
\label{sec:pixel_thresholds}
FIBAR operates {\em without} directly estimating each pixel's contrast threshold $\Coni$ and $\Coffi$.
This is partly driven by necessity because the popular line of Prophesee cameras used here does not provide
a frame readout of an intensity image, making direct calibration experimentally difficult. It also turns out that for many
image-related tasks, it is not necessary to reconstruct the brightness to scale, since for downstream processing, the
obtained image intensities are often rescaled anyway. Finally, operating without contrast threshold calibration
makes FIBAR significantly easier to use.

As outlined in Sec.~\ref{sec:method}, FIBAR estimates the imbalance between ON and OFF events for each pixel,
but what about the contrast threshold variance between pixels? Such fluctuations introduce
fixed pattern noise, and one could expect that correcting for it would be highly beneficial.

The purpose of this appendix is two-fold. For one, it establishes a framework and notation for how to reason about {\em relative}
contrast thresholds when {\em absolute} thresholds are not directly measurable. This allows a discussion of fixed
pattern threshold noise without using e.g.\ a Davis camera as was done in previous work\cite{brandli_2014}\cite{wang_ng_2019}.

The second part of this appendix shows that accounting for intra-pixel ON/OFF threshold imbalances the way FIBAR does, reduces
the impact of inter-pixel threshold variance on the reconstructed brightness. This partly explains why removing fixed pattern
threshold noise (Sec.~\ref{sec:inter_pixel_variation}) offered no observable benefits. 

\begin{figure}[t]
  \centering
  \includegraphics[width=1.0\linewidth]{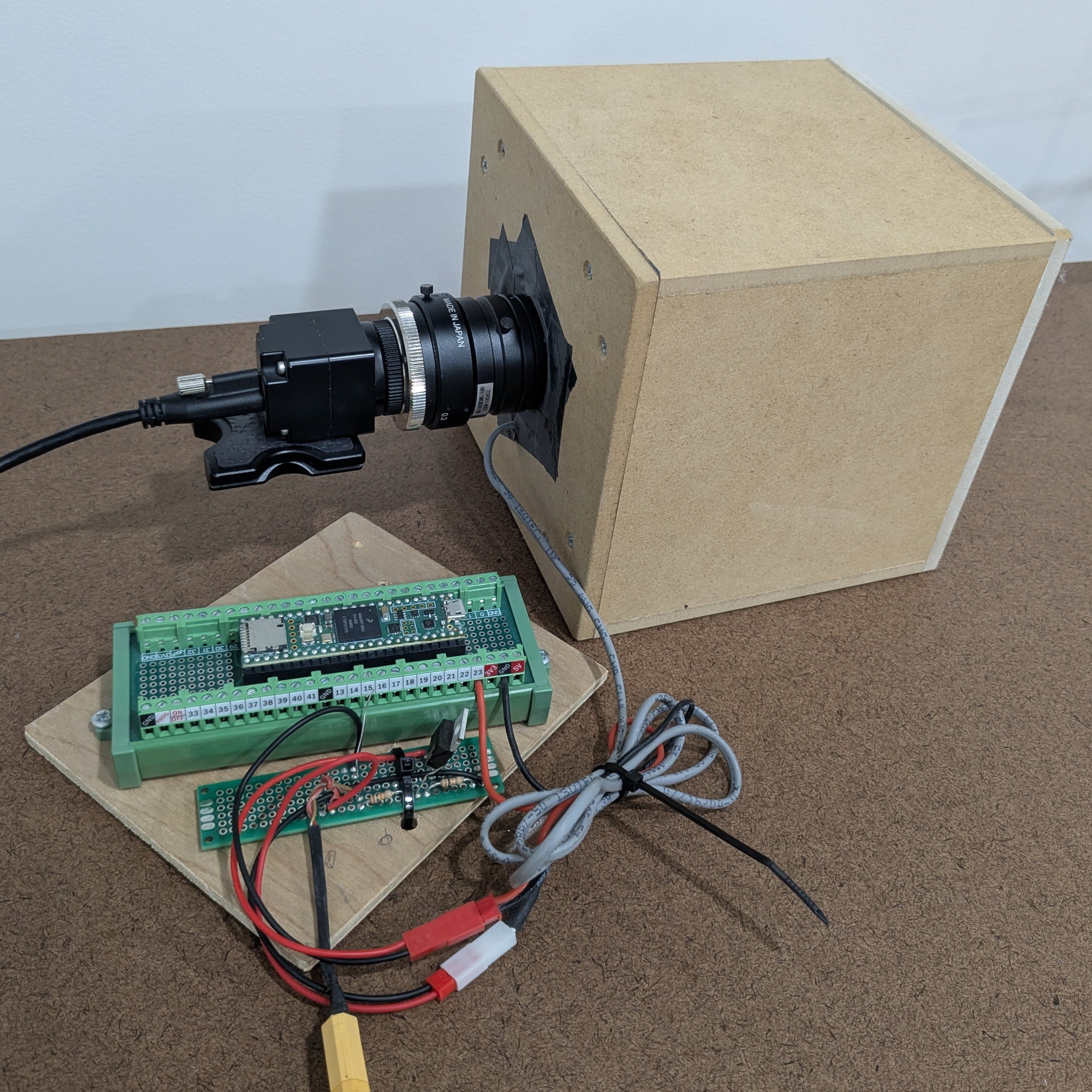}
  \caption{Experimental setup for measuring relative pixel thresholds.
  A SilkyEVCam Gen 3.1 640x480 (default biases) with an out-of-focus Kowa LM35HC lens is mounted
  against an illumination box (Fig.~\ref{fig:threshold_exp_inside}).
  The Teensy 4.1 (ARM Cortex-M7 at 600~MHz) controls an amplifier circuit based on an IRLB8721PBF MOSFET
  driving the LEDs inside the box with a 293~kHz PWM modulated at 1~Hz.}
  \label{fig:threshold_exp_overview}
\end{figure}
\begin{figure}[t]
  \centering
  \includegraphics[width=1.0\linewidth]{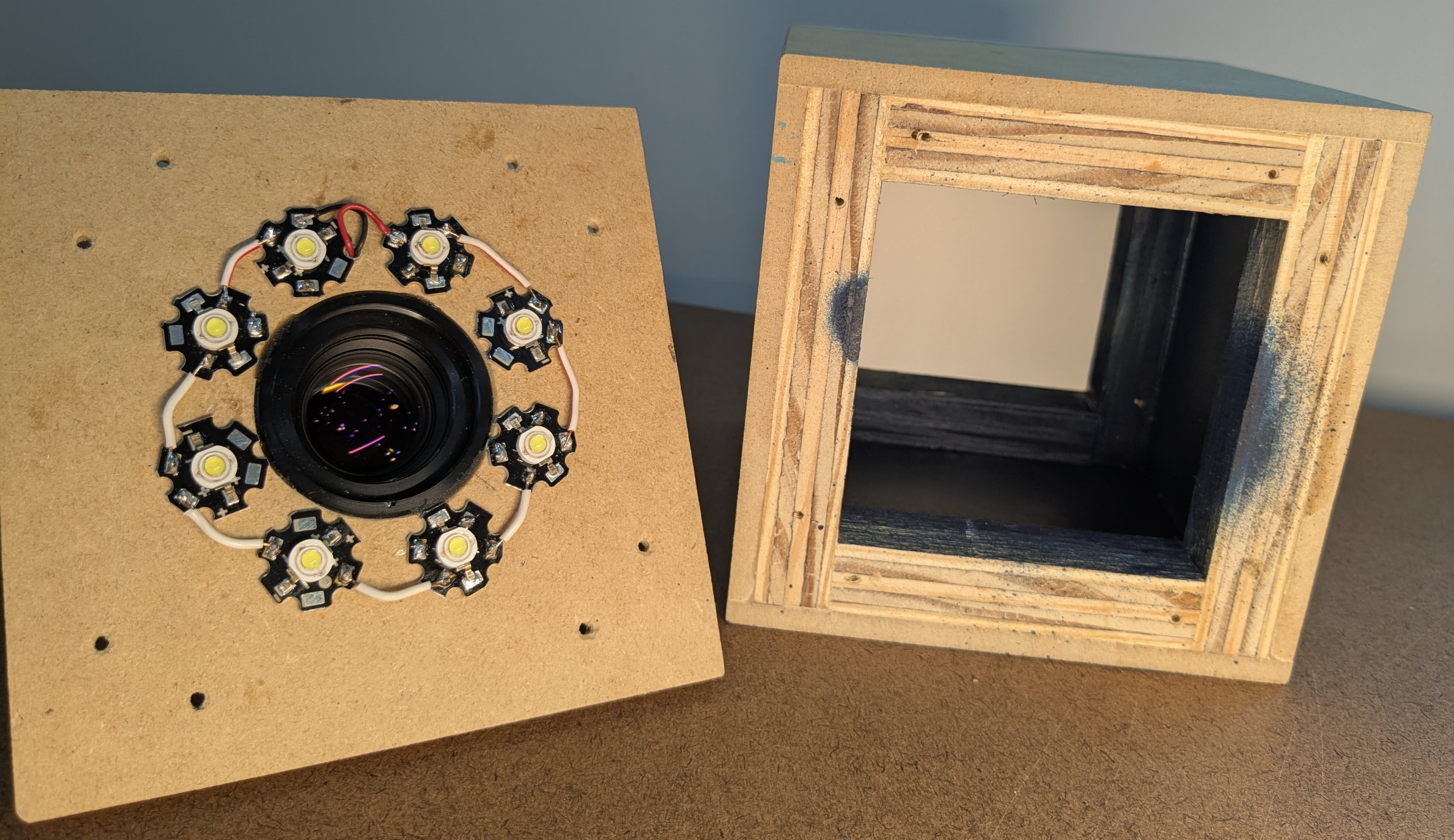}
  \caption{The back of the illumination box is painted uniformly white, the side walls are painted dark
  to obtain an adequately uniform illumination. Around the camera lens eight in-series SparkFun iPixel 3~W LEDs
  are mounted that are driven with a PWM signal of 24.5~V amplitude.}
  \label{fig:threshold_exp_inside}
\end{figure}

To measure relative thresholds (to be defined below), a camera (SilkyEVCam Gen 3.1, resolution 640x480, default biases)
with an out-of-focus 35~mm lens is mounted pointing at a uniformly illuminated screen that is enclosed
by a box to block out external light (Fig.~\ref{fig:threshold_exp_overview}).
Then the current for the LEDs illuminating the screen is ramped up and down {\em exponentially} at a frequency of 1~Hz
to yield a {\em linear} triangle wave for the brightness, see Eq.~(\ref{eq:definition_brightness}).
The low frequency of 1~Hz is chosen to avoid events being lost to sensor bandwidth saturation.
Note that the LED driver PWM frequency is 293~kHz and therefore does not trigger events since this is well beyond the cutoff
frequency of the low-pass filter of the sensor's front-end photo diode circuitry.
After recording for about 790 seconds, the relative thresholds can be estimated from the event data by leveraging
the stationarity of the signal as follows.

Calling $|\Delta L| = \sum_k|\Delta L_k|$  the total cumulative brightness {\em increase} at a given pixel due to all ON events in the recording, and
noting that the brightness does not drift from the signal mean during the experiment,
it follows that this is also to a good approximation the
total cumulative brightness {\em decrease}, yielding the following relationship between thresholds and the number of observed
events at pixel $i$:
\begin{equation}
    |\Delta L| = \Noni \Coni = \Noffi \Coffi .
\end{equation}
The total number of events $\Ntoti$ at pixel $i$ is then
\begin{equation}
    \Ntoti = \Noni + \Noffi = \frac{|\Delta L|}{\Coni} + \frac{|\Delta L|}{\Coffi} =  2\frac{|\Delta L|}{\Ci}\ ,
\end{equation}
where the threshold $\Ci$ is defined as the harmonic mean of $\Coni$ and $\Coffi$:
\begin{equation}
    \Ci := \frac{2 \Coni \Coffi}{\Coni + \Coffi}\ .
\end{equation}
Now, to compare the thresholds {\em between} pixels, a {\em global} threshold is defined that is common to all $\Npix$ sensor pixels.
In the context of the present experiment, where $\Delta L$ is roughly the same for all pixels, such
a threshold $C$ can be found as the harmonic mean of the individual pixel thresholds:
\begin{eqnarray}
    \Ntotbar&:=&\frac{1}{\Npix} \sum_i\Noni +\Noffi\\
    &=&\frac{1}{\Npix} \sum_i\frac{\absdeltaL}{\Coni} + \frac{\absdeltaL}{\Coffi}\\
    &=&2\absdeltaL\sum_i{\Ci}^{-1}/\Npix\\
    &=&2\absdeltaL / C
\end{eqnarray}
where
\begin{equation}
    C:=(\sum_i{\Ci}^{-1}/\Npix)^{-1}
    \label{eq:conip}
\end{equation}
can now be used to define a rescaled per-pixel threshold $\Cip = \Ci / C$ with a harmonic average of 1.
Defining relative thresholds this way makes it easy to compute them directly from the observed event counts, e.g.:
\begin{eqnarray}
  \Conip&=&\Coni / C = \frac{1}{2}\Ntotbar / \Noni\\
  \Cip&=&\Ntotbar / (\Noni + \Noffi)\ .\label{eq:cip}
\end{eqnarray}

\begin{figure}[t]
  \centering
  \includegraphics[width=1.0\linewidth]{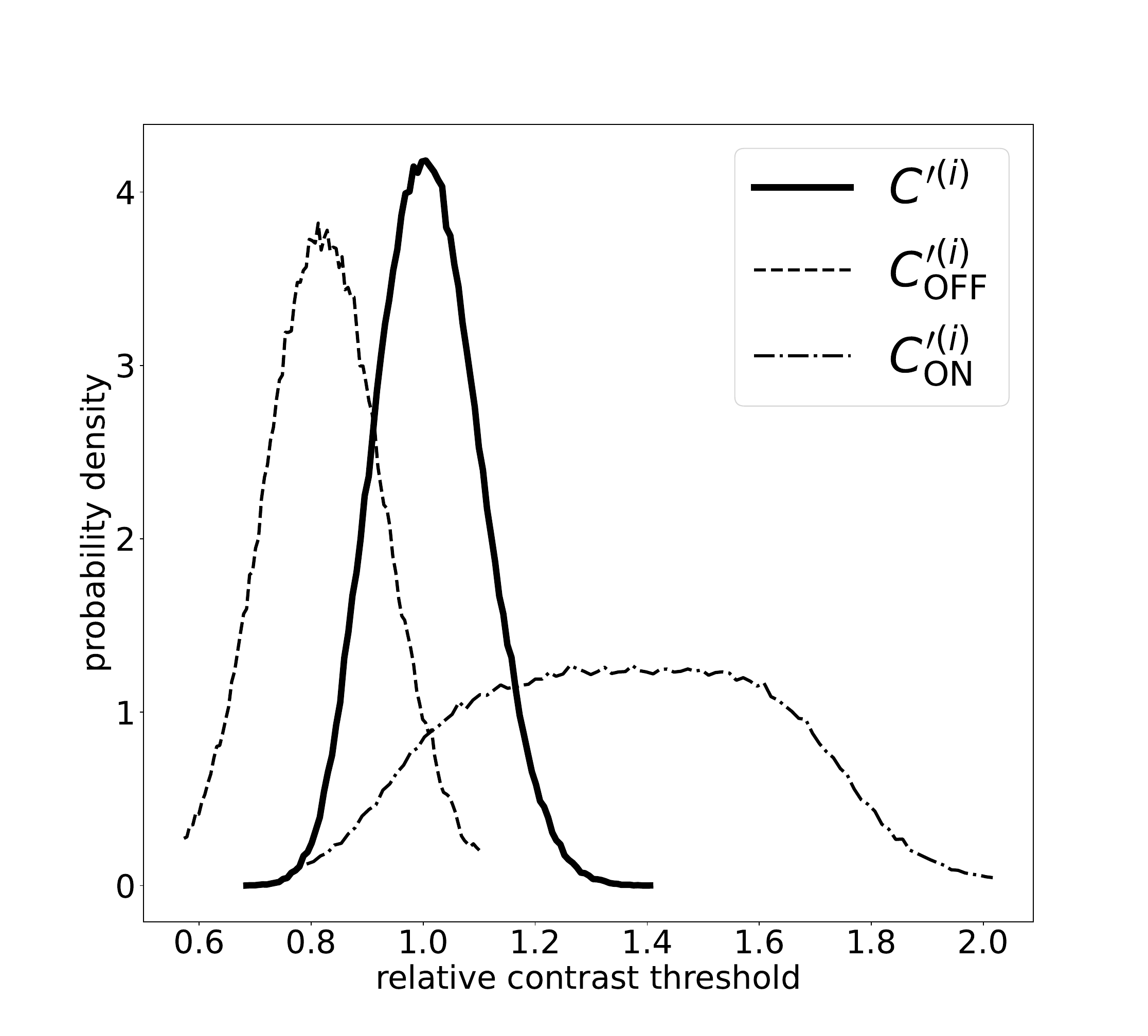}
  \caption{Experimentally observed relative contrast thresholds $C^\prime$, computed analogously to Eq.~(\ref{eq:conip}).
  The top 1\% highest and lowest count pixels have been removed for readability.
  }
  \label{fig:scaled_thresholds}
\end{figure}
As can be seen from Fig.~\ref{fig:scaled_thresholds}, the distributions of $\Conip$ and $\Coffip$ are much
broader than the one for $\Cip$. This means that by compensating for the ON/OFF threshold imbalance, FIBAR
captures the most important aspect of inter-pixel threshold variance.
This explains at least partially why correcting for inter-pixel threshold variation in Sec.~\ref{sec:inter_pixel_variation}
yielded no improvement.

\section{Acknowledgements}
The author acknowledges useful and encouraging discussions with
Kenneth Chaney, Fernando Cladera, Kostas Daniilidis, Ziyun Wang, and Andreas Ziegler.

{\small
\bibliographystyle{ieee_fullname}
\bibliography{fibar}

\begin{thebibliography}{10}\itemsep=-1pt

\bibitem{brandli_2014}
Christian Brandli, Lorenz Muller, and Tobi Delbruck.
\newblock Real-time, high-speed video decompression using a frame- and event-based {DAVIS} sensor.
\newblock In {\em 2014 IEEE International Symposium on Circuits and Systems (ISCAS)}, pages 686--689, 2014.

\bibitem{chakravarthi2024recenteventcamerainnovations}
Bharatesh Chakravarthi, Aayush~Atul Verma, Kostas Daniilidis, Cornelia Fermuller, and Yezhou Yang.
\newblock Recent event camera innovations: A survey, 2024.

\bibitem{chaney_m3ed}
Kenneth Chaney, Fernando Cladera, Ziyun Wang, Anthony Bisulco, M.~Ani Hsieh, Christopher Korpela, Vijay Kumar, Camillo~J. Taylor, and Kostas Daniilidis.
\newblock {M3ED}: Multi-robot, multi-sensor, multi-environment event dataset.
\newblock In {\em 2023 IEEE/CVF Conference on Computer Vision and Pattern Recognition Workshops (CVPRW)}, pages 4016--4023, 2023.

\bibitem{dong2025eventbasedfastintensityreconstruction}
Xin Dong, Yiwei Zhang, Yangjie Cui, Jinwu Xiang, Daochun Li, and Zhan Tu.
\newblock An event-based fast intensity reconstruction scheme for {UAV} real-time perception, 2025.

\bibitem{hypere2vid2024}
Burak Ercan, Onur Eker, Canberk Saglam, Aykut Erdem, and Erkut Erdem.
\newblock {HyperE2VID}: Improving event-based video reconstruction via hypernetworks.
\newblock {\em Trans. Img. Proc.}, 33:1826–1837, Mar. 2024.

\bibitem{kalibr_2013}
Paul Furgale, Joern Rehder, and Roland Siegwart.
\newblock Unified temporal and spatial calibration for multi-sensor systems.
\newblock In {\em 2013 IEEE/RSJ International Conference on Intelligent Robots and Systems}, pages 1280--1286, 2013.

\bibitem{gallego_delbrueck_orchard}
Guillermo Gallego, Tobi Delbrück, Garrick Orchard, Chiara Bartolozzi, Brian Taba, Andrea Censi, Stefan Leutenegger, Andrew~J. Davison, Jörg Conradt, Kostas Daniilidis, and Davide Scaramuzza.
\newblock Event-based vision: A survey.
\newblock {\em IEEE Transactions on Pattern Analysis and Machine Intelligence}, 44(1):154--180, 2022.

\bibitem{shining_light_2023}
Rui Graça, Brian McReynolds, and Tobi Delbruck.
\newblock Shining light on the {DVS} pixel: A tutorial and discussion about biasing and optimization.
\newblock In {\em 2023 IEEE/CVF Conference on Computer Vision and Pattern Recognition Workshops (CVPRW)}, pages 4045--4053, 2023.

\bibitem{mostafavi2020}
S.~Mohammad~Mostafavi I., Lin Wang, Yo{-}Sung Ho, and Kuk{-}Jin Yoon.
\newblock Event-based high dynamic range image and very high frame rate video generation using conditional generative adversarial networks.
\newblock {\em CoRR}, abs/1811.08230, 2018.

\bibitem{libcaer}
{iniVation AG}.
\newblock {libcaer}: Minimal {C} library to access, configure and get data from neuromorphic sensors and processors.
\newblock \url{https://gitlab.com/inivation/dv/libcaer}, 2014--2025.
\newblock Accessed: 2025-09-02.

\bibitem{lichtsteiner_posch_delbrueck_2008}
Patrick Lichtsteiner, Christoph Posch, and Tobi Delbruck.
\newblock A 128$\times$ 128 120 db 15 $\mu$s latency asynchronous temporal contrast vision sensor.
\newblock {\em IEEE Journal of Solid-State Circuits}, 43(2):566--576, 2008.

\bibitem{fedi_2023}
Shijie Lin, Yingqiang Zhang, Dongyue Huang, Bin Zhou, Xiaowei Luo, and Jia Pan.
\newblock Fast event-based double integral for real-time robotics.
\newblock In {\em 2023 IEEE International Conference on Robotics and Automation (ICRA)}, pages 796--803, 2023.

\bibitem{manderscheid_2019_cvpr}
Jacques Manderscheid, Amos Sironi, Nicolas Bourdis, Davide Migliore, and Vincent Lepetit.
\newblock Speed invariant time surface for learning to detect corner points with event-based cameras.
\newblock In {\em Proceedings of the IEEE/CVF Conference on Computer Vision and Pattern Recognition (CVPR)}, June 2019.

\bibitem{muglikar_calib_2021}
Manasi Muglikar, Mathias Gehrig, Daniel Gehrig, and Davide Scaramuzza.
\newblock How to calibrate your event camera.
\newblock In {\em Proceedings of the IEEE/CVF Conference on Computer Vision and Pattern Recognition (CVPR) Workshops}, pages 1403--1409, June 2021.

\bibitem{munda_2018}
Gottfried Munda, Christian Reinbacher, and Thomas Pock.
\newblock Real-time intensity-image reconstruction for event cameras using manifold regularisation.
\newblock {\em Int. J. Comput. Vision}, 126(12):1381–1393, Dec. 2018.

\bibitem{orchard2021loihi2}
Garrick Orchard, E.~Paxon Frady, Daniel Ben~Dayan Rubin, Sophia Sanborn, Sumit~Bam Shrestha, Friedrich~T. Sommer, and Mike Davies.
\newblock Efficient neuromorphic signal processing with {Loihi} 2.
\newblock In {\em 2021 IEEE Workshop on Signal Processing Systems (SiPS)}, pages 254--259, 2021.

\bibitem{paredes_2021}
Federico Paredes-Vallés and Guido C. H.~E. de Croon.
\newblock Back to event basics: Self-supervised learning of image reconstruction for event cameras via photometric constancy.
\newblock In {\em 2021 IEEE/CVF Conference on Computer Vision and Pattern Recognition (CVPR)}, pages 3445--3454, 2021.

\bibitem{pfrommer_2022_frequencycam}
Bernd Pfrommer.
\newblock {Frequency Cam}: Imaging periodic signals in real-time.
\newblock 2022.
\newblock arXiv, https://arxiv.org/abs/2211.00198.

\bibitem{proakis_manolakis_2006}
John~G. Proakis and Dimitris~K Manolakis.
\newblock {\em Digital Signal Processing (4th Edition)}.
\newblock Prentice Hall, 4 edition, 2006.

\bibitem{RebecqE2VID2019}
Henri Rebecq, Ren{\'{e}} Ranftl, Vladlen Koltun, and Davide Scaramuzza.
\newblock High speed and high dynamic range video with an event camera.
\newblock {\em {IEEE} Trans. Pattern Anal. Mach. Intell. (T-PAMI)}, 2019.

\bibitem{rosa_2022}
Leandro de~Souza Rosa, Aiko Dinale, Simeon Bamford, Chiara Bartolozzi, and Arren Glover.
\newblock High-throughput asynchronous convolutions for high-resolution event-cameras.
\newblock In {\em 2022 8th International Conference on Event-Based Control, Communication, and Signal Processing (EBCCSP)}, pages 1--8, 2022.

\bibitem{salah_ecalib_2024}
Mohammed Salah, Abdulla Ayyad, Muhammad Humais, Daniel Gehrig, Abdelqader Abusafieh, Lakmal Seneviratne, Davide Scaramuzza, and Yahya Zweiri.
\newblock {E-Calib}: A fast, robust, and accurate calibration toolbox for event cameras.
\newblock {\em IEEE Transactions on Image Processing}, 33:3977--3990, 2024.

\bibitem{scheerlinck2019asynchronousconvolutions}
Cedric Scheerlinck, Nick Barnes, and Robert Mahony.
\newblock Asynchronous spatial image convolutions for event cameras.
\newblock {\em IEEE Robotics and Automation Letters}, 4(2):816--822, 2019.

\bibitem{scheerlinckcontinuous2018}
Cedric Scheerlinck, Nick Barnes, and Robert Mahony.
\newblock Continuous-time intensity estimation using event cameras.
\newblock In C.V. Jawahar, Hongdong Li, Greg Mori, and Konrad Schindler, editors, {\em Computer Vision -- ACCV 2018}, pages 308--324, Cham, 2019. Springer International Publishing.

\bibitem{scheerlink2020firenet}
Cedric Scheerlinck, Henri Rebecq, Daniel Gehrig, Nick Barnes, Robert~E. Mahony, and Davide Scaramuzza.
\newblock Fast image reconstruction with an event camera.
\newblock In {\em 2020 IEEE Winter Conference on Applications of Computer Vision (WACV)}, pages 156--163, 2020.

\bibitem{stoffregen_2020}
Timo Stoffregen, Cedric Scheerlinck, Davide Scaramuzza, Tom Drummond, Nick Barnes, Lindsay Kleeman, and Robert Mahony.
\newblock Reducing the sim-to-real gap for event cameras.
\newblock In {\em Computer Vision – ECCV 2020: 16th European Conference, Glasgow, UK, August 23–28, 2020, Proceedings, Part XXVII}, page 534–549, Berlin, Heidelberg, 2020. Springer-Verlag.

\bibitem{wang_apriltags_2016}
John Wang and Edwin Olson.
\newblock Apriltag 2: Efficient and robust fiducial detection.
\newblock In {\em 2016 IEEE/RSJ International Conference on Intelligent Robots and Systems (IROS)}, pages 4193--4198, 2016.

\bibitem{wang_gan_2019}
Lin Wang, I.S.~Mohammad Mostafavi, Yo-Sung Ho, and Kuk-Jin Yoon.
\newblock Event-based high dynamic range image and very high frame rate video generation using conditional generative adversarial networks.
\newblock In {\em 2019 IEEE/CVF Conference on Computer Vision and Pattern Recognition (CVPR)}, pages 10073--10082, 2019.

\bibitem{wang_ng_2019}
Ziwei Wang, Yonhon Ng, Pieter van Goor, and Robert Mahony.
\newblock Event camera calibration of per-pixel biased contrast threshold.
\newblock In {\em Australasian Conference of Robotics and Automation (ACRA)}, 2019.

\bibitem{alex_zhu_tensor_2019}
Alex~Zihao Zhu, Liangzhe Yuan, Kenneth Chaney, and Kostas Daniilidis.
\newblock Live demonstration: Unsupervised event-based learning of optical flow, depth and egomotion.
\newblock In {\em 2019 IEEE/CVF Conference on Computer Vision and Pattern Recognition Workshops (CVPRW)}, pages 1694--1694, 2019.

\bibitem{eventhdr_2024}
Yunhao Zou, Ying Fu, Tsuyoshi Takatani, and Yinqiang Zheng.
\newblock Eventhdr: From event to high-speed {HDR} videos and beyond.
\newblock {\em IEEE Transactions on Pattern Analysis and Machine Intelligence}, 47(1):32--50, 2025.

\end{thebibliography}
}

\end{document}